\pdfoutput=1

\documentclass[11pt]{article}

\usepackage[]{ACL2023}

\usepackage{times}
\usepackage{latexsym}

\usepackage[T1]{fontenc}

\usepackage[utf8]{inputenc}

\usepackage{microtype}

\usepackage{inconsolata}
\usepackage{booktabs}
\usepackage{graphicx} 
\usepackage{csquotes}
\usepackage{dblfloatfix}

\usepackage{xspace}
\newcommand\approachname{\textsc{L\&O}\xspace}
\newcommand\better{\textsc{Better}\xspace}
\newcommand\basic{Basic\xspace}

\title{Granting GPT-4 License and Opportunity: Enhancing Accuracy and Confidence Estimation for Few-Shot Event Detection}

\author{Steven Fincke \\ University of Southern California \\ Information Sciences Institute \\ sfincke@isi.edu 
        \AND
        Adrien Bibal \\ InferLink \\ abibal@inferlink.com  \And
        Elizabeth Boschee \\ University of Southern California \\ Information Sciences Institute \\ boschee@isi.edu }

\begin{document}
\maketitle
\begin{abstract}
Large Language Models (LLMs) such as GPT-4 have shown enough promise in the few-shot learning context to suggest use in the generation of "silver" data and refinement of new ontologies through iterative application and review. Such workflows become more effective with reliable confidence estimation. Unfortunately, confidence estimation is a documented weakness of models such as GPT-4, and established methods to compensate require significant additional complexity and computation. 
The present effort explores methods for effective confidence estimation with GPT-4
with few-shot learning for event detection in the \better ontology as a vehicle.
The key innovation is expanding the prompt and task presented to GPT-4 
to provide \textbf{L}icense to speculate when unsure \textbf{a}nd 
\textbf{O}pportunity to quantify and explain its uncertainty (\approachname).
This approach improves accuracy and provides usable confidence measures (0.759 AUC) with no additional machinery.
\end{abstract}

\section{Introduction}
Large language models such as GPT-4 have shown particular utility in the few-shot learning context,
offering the promise of facilitating the creation of large annotation sets for new tasks.
This study explores prompting strategies to enhance confidence estimation, which we refer to as \textit{\textbf{L}icense \& \textbf{O}pportunity} (\approachname). 
Our task is the detection of events in English-language news stories in the \better ontology \cite{mckinnon-rubino-2022-iarpa}.
\approachname simply uses one query to the LLM to obtain the output along with the confidence estimation, and it does not require access to the LLMs internal statistics, nor any LLM fine-tuning. 
Reliable confidence estimation is particularly useful when such a tool is used for tasks with categories that are easily confused, leading to low inter-rater agreement. 
It also guides the evaluation of LLM output, both for purposes of refining an ontology under development and prioritizing the review of "silver" annotations.

\begin{table*}[t]
\centering
\begin{tabular}{p{0.1\linewidth} | p{0.2\linewidth} | p{0.6\linewidth}}
\toprule
              Topic                   & Description & Event types           \\ \midrule
Law & Law enforcement and judicial actions  
& Law-Enforcement-Arrest, Law-Enforcement-Other, Judicial-Indict, Judicial-Prosecute, Judicial-Convict, Judicial-Sentence, Judicial-Acquit, Judicial-Seize, Judicial-Plead, Judicial-Other \\ \midrule
Violence                 & Associated with military action and terrorism 
& Violence, Violence-Attack, Violence-Bombing, Violence-Kill, Violence-Wound, Violence-Damage, Violence-Other Kidnapping \\ \midrule
Disease                  & Disease outbreaks and responses 
& Impose-Quarantine, Apply-NPI, Hospitalize, Vaccinate, Test-Patient, Treat-Patient, Conduct-Medical-Research, Disease-Outbreak, Disease-Infects, Disease-Exposes, Disease-Kills, Disease-Recovery, Restrict-Business \\ \bottomrule
\end{tabular}
\caption{Selected topics within the \better \basic ontology \cite{mckinnon-rubino-2022-iarpa}. (\textit{NPI}: "non-pharmaceutical intervention")}
\label{tab:topics}
\end{table*}

\section{Related Work}
Few-shot learning has been successfully applied with neural LMs, including LLMs, to the task of event detection \cite{barth2022few, gao2024eventrl}; i.e., the detection of event phrases (\textit{triggers} or \textit{anchors}) and labeling the according to a specified ontology of event types. Such methods have even been applied to highly inclusive event ontologies such as Open Information Extraction \cite{ling2023improving, mccusker2023loke}. 
\citet{deng2023information} provide a survey of recent work in information extraction and directly compare various approaches built upon BERT-like LMs and those using LLMs;
they report 0.539 micro F1 for the full ACE05 event detection task
using a 5-shot training strategy with GPT-4,
where the SOTA fine-tuning a BERT-like LM with a full training set is at 0.837 micro F1.


Confidence estimation, however, has not been a focus of prior scholarship on few-shot event detection with LLMs. 
If we consider a broader array of NLP tasks,
we see that various methods have been developed to extract reliable confidence estimations from LLMs. 
The fact that GPT-4 usually produces high confidence values when asked naively significantly complicates the task \cite{singh2023confidencecompetence}.
Many studies frame the problem as \textit{confidence calibration}; that is calibrating the output confidence probability to the actual observed probability of correctness in a labeled dataset \cite{guo2017calibration, tian-etal-2023-just}. One family of approaches exploits the internal statistics of the LLM, such as the log-probs of tokens conveying a particular answer. For example, \citet{wu2024reliable} compare variants of "Einstein was born in the year \textit{X}", where \textit{X} is \textit{1878}, \textit{1879}, or \textit{1880}. They observe the log probabilities for the various year strings, hopefully providing the highest probability for \textit{1879}.
Such \textit{white box} approaches contrast with \textit{black box} techniques which do not use such internal statistics. 
Some find an advantage in using natural language expressions of confidence instead of generating numbers \cite{lin2022teaching, tian-etal-2023-just}.
These techniques, e.g. \citet{singh2023confidencecompetence}, often involve complex, multi-stage strategies such as "Chain of Thought" \cite{wei2023chainofthought} and "Tree of Thoughts" \cite{yao2023tree}.

One important feature of \approachname (our approach) is requesting explanations in addition to answers and confidence ratings. Some pre-existing "black-box" methods also include the generation of explanations as part of their confidence pipelines.
\citet{li2024think} generate justifications for each of a few possible answers and then estimates confidence from these explanations. 
\citet{xiong2024llms} report that their \textit{self probing} prompting method is particularly effective for GPT-4; this approach considers each possible answer to a question separately and requests an explanation and confidence; these are reviewed together to generate normalized confidence levels for all the options. 
Unlike \approachname, these do not provide explanations along with the final confidence output.

\section{Data and Task}
We utilize here a portion of IARPA's \better task \cite{mckinnon-rubino-2022-iarpa}, which focuses on information extraction from news stories in the cross-lingual context. 
The \basic portion has an event ontology which eventually expanded, in the third and final phase of the program, to 114 categories; 
these are grouped into 12 topics ranging from crime to finance. All the annotations for development are for English texts. The full three phases provided annotations for 732 English-language news stories.
However, the program evaluations only considered performance on texts in Arabic, Farsi, Russian, Chinese, and Korean. 

An effective solution to the \better \basic event extraction task uses BIO token labeling to mark event phrases for the full ontology in one pass \cite{jenkins-etal-2023-massively}. XLM-RoBERTa-large \cite{conneau-etal-2020-unsupervised} was fine-tuned to the full training set with the base model providing effective zero-shot cross-lingual transfer from the English training data to the \better program languages. 

By contrast, this effort utilizes few-shot learning with a single submission to an LLM. We eliminate the cross-lingual aspect, evaluating on English data, instead. We ask the LLM to mark only the beginning of each event phrase with a vertical pipe (|); we do not ask for the entire span because of the difficulty of crafting effective guidelines for selecting the exact scope of phrases in the few-shot context.
For resource reasons, 
we constrain our efforts to only three of the twelve \textit{basic} topics, as detailed in Table \ref{tab:topics}. We also exclude all event types with fewer than 10 instances in all our available annotations. 

\section{System}
\begin{figure*}[!h]
\centering
\includegraphics[width=16cm]{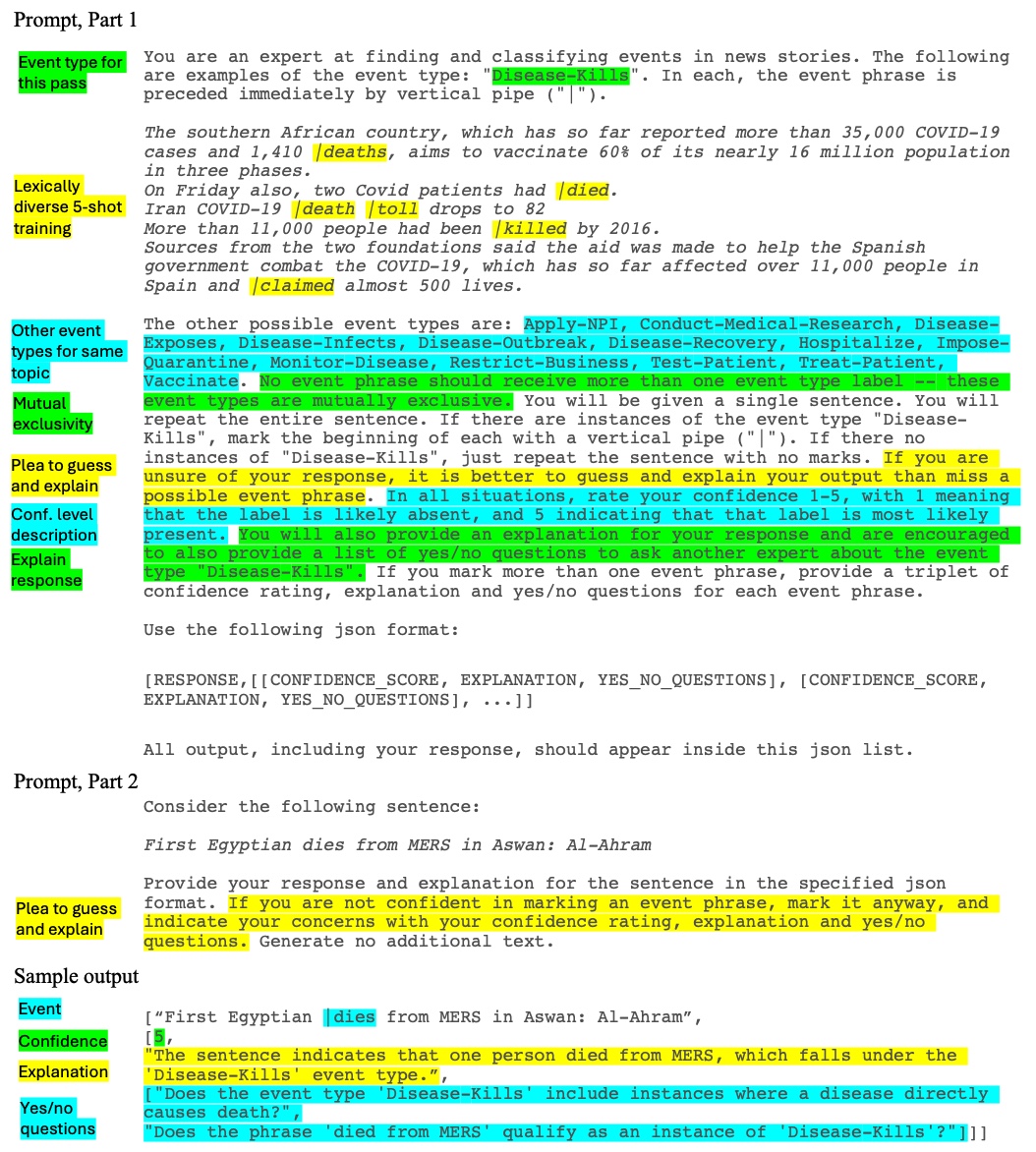}
\caption{Sample prompt and output for \textit{Disease-Kills} within the \textit{Disease} topic.}
\label{fig:prompt}
\end{figure*}

We provide example prompts and raw output in Figure \ref{fig:prompt}. 
We will first highlight some important features of our approach and then explain details. 
\approachname adopts two related strategies: 1) urging the LLM to provide guesses when in doubt, and 2) providing the LLM ample opportunity to characterize uncertainty in generating the event type label (or lack thereof). 
The latter was provided, in part, to allow consumers of the output to differentiate between outright guesses and confident answers.

The LLM is prompted to provide a confidence rating ranging 1-5, where 5 indicates the highest confidence in the \textbf{presence} of an event of the specified type in the sentence, and 1 marks the greatest confidence in the \textbf{absence} of the event type. 
1 is appropriate when no event instances are predicted, as well as for very unlikely guesses. 
Unlike more conventional approaches, this does not convey the LLM's confidence in its answer, regardless of its content; the consequences of such a strategy are discussed in Section \ref{ablations}.

A complete response also includes an explanation. Additionally, the LLM is encouraged to ask a (fictional) expert yes/no questions about the specific event type, with an eye on refining annotation conventions.
These requests are inspired by techniques such as CoT \cite{wei2023chainofthought},
but our prompt provides no example of expected explanations or yes/no questions and does not encourage decomposing the task into simpler logical steps.
Though not explicitly encouraged by our prompt, the LLM often uses the explanation to account for its confidence rating, not just the presence or absence of the label. 

\subsection{Task and prompting details}
Each topic is treated as a separate task, and each query to the LLM asks for labels for a single event type at a time, but the LLM is told which other event types fall within the same topic. 
This design simplifies the task by allowing the LLM to focus on one event type at a time;
it increases the possibility of accuracy suffering due to incompatible outputs for different event types for the same sentence, but the scope of this effort does not include a mechanism for reconciling output for related event types.

Our prompt provides up to 5 sentences with at least one instance of the event type in question marked with a vertical pipe before the first word of each event phrase. To select these, we group each instance of the given event type by the lemma of the first event word, ignoring those with only one instance, and then randomly sample according to lexical type. If fewer than five lexical types are attested, more examples are taken from more frequent types. All few-shot examples are excluded from the testing pool, as well as all sentences with fewer than 25 characters. The test repeats a cycle of one sentence with at least one event in the chosen topic in the reference annotations and then three sentences with no in-topic events. The prompt lists the names of other event types in the same topic and notes that no word can bear more than one event type label, i.e., they are mutually exclusive. 

The LLM provides its answer by repeating the sentence with vertical pipes marking the beginning of event phrases; our scripts include text alignment code to be robust to the imperfections in the LLM's copy. If there are no instances of the event in the sentence, the LLM is simply instructed to repeat the whole sentence unaltered. As discussed earlier, the LLM is also asked to provide a confidence rating, explanation and a list of yes/no questions. When more than one event phrase is indicated, the LLM generates a separate \textit{triplet} for each event phrase. 

\subsection{Scoring}
We evaluate the LLM's predictions according to the output confidence level.
The explanation and yes/no questions are not used by any subsequent step \citep[unlike][]{li2024think, xiong2024llms} and not in adjudication,
but Section \ref{ablations} will demonstrate their contribution to system performance. 
A response is judged correct if the marked word coincides with the first word of a reference event phrase of the specified type. No credit is given for marking another word within a phrase or indicating a related but different event type. If the reference has multiple phrases for the same event, the LLM is expected to mark the beginning of each. 
As we sweep our confidence levels,
we include all the outputs for \textbf{higher} levels of confidence (if any). 
We characterize the performance for each topic with precision, recall, and macro F1 score of the confidence level with the highest F1. We also compute a ROC AUC \footnote{\textit{roc\_auc\_score} from \textit{sklearn}}:
for each positive output from our system, we provide the generated confidence score
and a label for correctness.
This statistic indicates the probability of a randomly selected true positive having a higher confidence score than a randomly selected false positive. 
\begin{table}[t]
\centering
\setlength\tabcolsep{5pt}
\begin{tabular}{@{}lllll@{}}
\toprule
Topic & P & R & F1 & AUC \\ \midrule
Law & 0.466 & 0.547 & 0.503 & 0.774 \\ \midrule
Violence & 0.432 & 0.451 & 0.441 & 0.729 \\ \midrule
Disease & 0.491 & 0.396 & 0.439 & 0.645 \\ \bottomrule
\end{tabular}
\caption{Precision, recall, F1 and AUC for the top-performing confidence level for selected topics.}
\label{tab:topic_performance}
\end{table} 
Table \ref{tab:topic_performance} presents performance for three selected topics: \textit{Law}, \textit{Violence}, and \textit{Disease}.
We have no external baseline for our version of the Basic task, but we can provide some related reference points. The SOTA for event detection from English text for the full \better \basic task ranges 0.642-0.646 F1, varying according to phase variants of the task \cite{jenkins-etal-2023-massively}. 
The ACE05 event detection has been well explored; on average, it is somewhat easier than \better \basic; \citet{jenkins-etal-2023-massively} report 0.712 F1 for English ACE05. 
\citet{deng2023information} cite 0.837 F1 as the SOTA for the full English ACE05, but 0.539 F1 with GPT-4 employing a few-shot strategy.
This prior work leads us to regard the performance of \approachname as reasonably accurate. 

\begin{figure}[!h] 
\centering
\includegraphics[width=1.0\columnwidth]{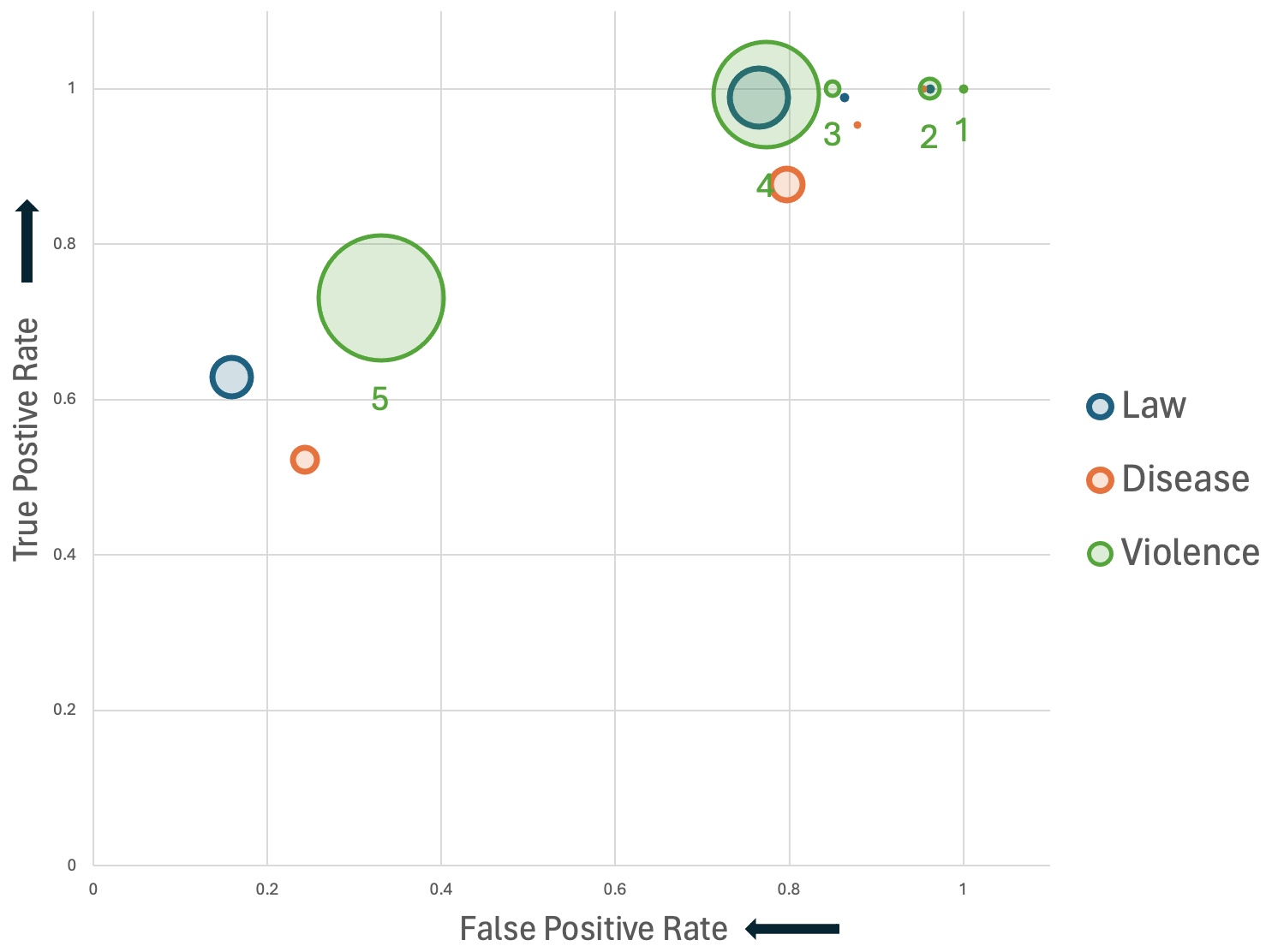}
\caption{AUC plot for three topics with diameter proportional to the number of outputs at the specified confidence level.}
\label{fig:roc_three_topics}
\end{figure}

To facilitate the interpretation of the AUC scores,
we include 
an ROC-style 
plot of the performance for the three topics:
Figure \ref{fig:roc_three_topics};
The portions of true positives and false positives are calculated in the same manner as Table \ref{tab:topic_performance}: 
i.e., lower levels of confidence include all predictions with higher confidence. 
(Recall that the best performance is towards the top-left corner of such graphs.) Please note, however, that the diameter of each plot point is proportional to the number of instances at an indicated confidence level.
This plot shows that most predictions have a confidence score of 5 or 4, 
with significant additions of both true and false positives when including 4. 
The values for 5 and 4 are remarkably similar, 
indicating that GPT-4 was consistent in determining the precision/recall trade off between 5 and 4 in the separate runs. 
All three have AUC values much greater than chance at 0.5; indeed, \textit{Law} and \textit{Violence} are above 0.7. 

\section{Ablation studies} \label{ablations}
We will explore here the impact of various design features of \approachname. To contain this effort, we will focus on the initial 210 sentences of the run for the \textit{Law} topic. 
In one set of variants, we exclude components of \approachname 
that are means for the LLM to characterize its stance to its output. 
First, we request a confidence level for each output but do not ask for an explanation or any yes/no questions.
We add a variant of this where the confidence score is more \textit{conventional}; i.e., the prompt asks the LLM to generate a 5 if it is highly confident in its answer and 1 if quite uncertain \textbf{regardless} of the content of the answer; a 5 score can be provided when highly confident in the absence, as well as presence, of an event. 
We also consider the inverse, where the prompt requests both an explanation and list of yes/no questions but no confidence score. 
We exclude our \approachname pleas for guessing with explanation in the last variant.
Table \ref{tab:law_210_accounting_ablations} indicates performance on our 210-sentence \textit{Law} subset for our full system and these ablations.

\begin{table}[h!]
\centering
\setlength\tabcolsep{5pt}
\begin{tabular}{@{}lllll@{}}
\toprule
Accounting & P & R & F1 & AUC \\ \midrule
Full & 0.543 & 0.667 & 0.599 & 0.759 \\ \midrule
Conf. only & 0.775 & 0.378 & 0.508 & 0.818 \\ \midrule
Conv. conf. only & 0.350 & 0.448 & 0.393 & 0.652 \\ \midrule
No conf. & 0.417 & 0.577 & 0.484 & 0.500 \\  \midrule
No guess & 0.540 & 0.627 & 0.580 & 0.580 \\ \bottomrule
\end{tabular}
\caption{Baseline for the \textit{Law} topic with ablations excluding explanation and yes/no questions.}
\label{tab:law_210_accounting_ablations}
\end{table}

Requesting confidence only, i.e., dropping the explanation and yes/no questions, degrades F1 somewhat, especially for recall, but the AUC is actually somewhat better. 
Since explanations often provide rationales for the confidence rating,
not just the labeling decision,
it appears that GPT-4 generates better confidence rating when given the opportunity to account for its ouptut.
However, the \textit{conventional} variant, where 5 is requested when the LLM is highly confident in both positive labels and the lack thereof, is markedly worse in terms of F1 and AUC. 
Excluding the confidence ratings necessarily reduces the AUC to chance, but the F1 is comparable to \textit{Confidence only}.
Quite strikingly, dropping the appeals to guess leaves F1 largely unchanged in relation to the full system, except for a modest drop in recall, but the confidence scores lose most of their value, and the AUC is only 0.080 above chance. 

\begin{figure}[h]
\centering
\includegraphics[width=1.0\columnwidth]{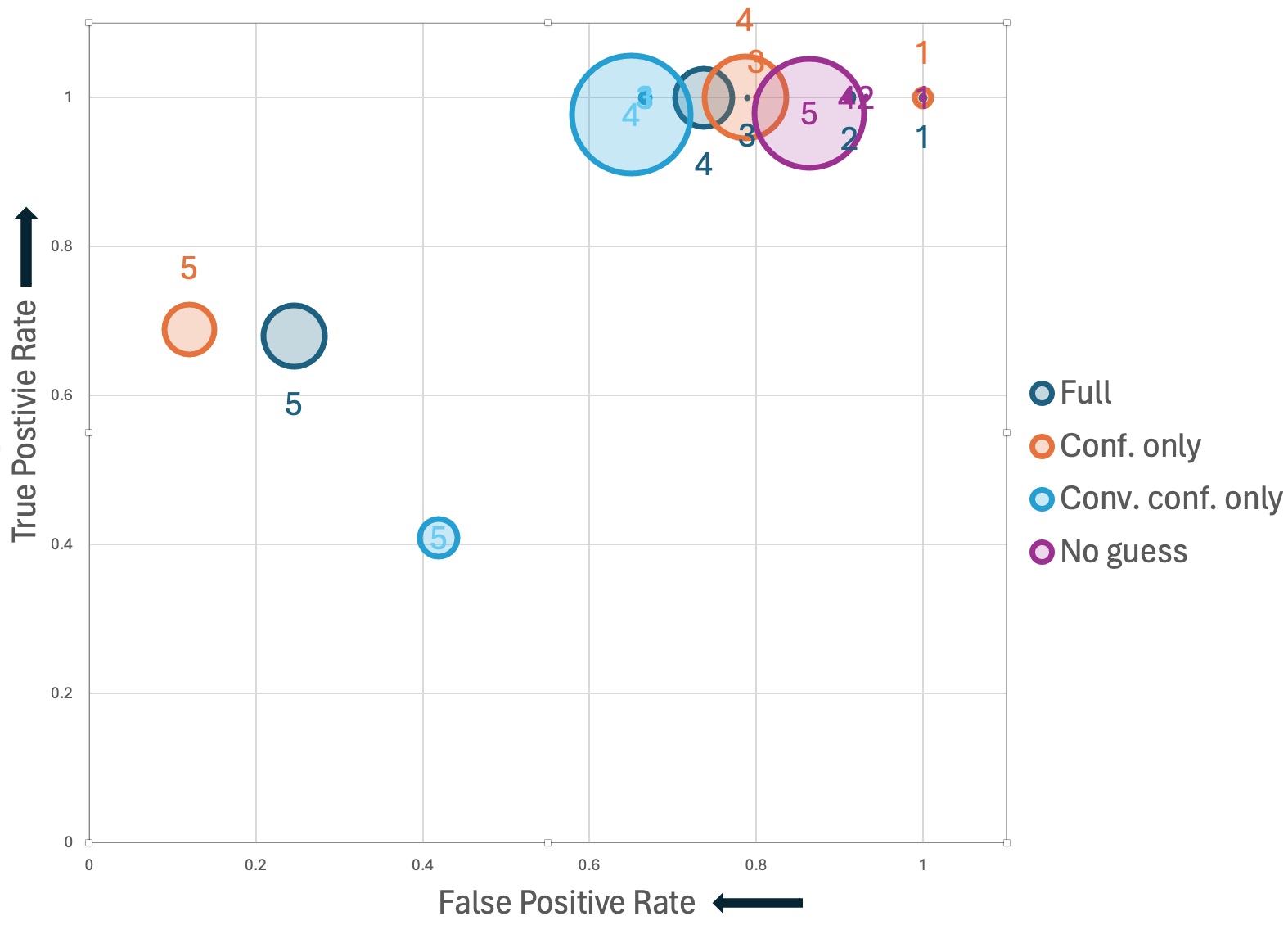}
\caption{
AUC plot for individual confidence levels for our full system and ablation variants. The diameter proportional to the number of outputs at the specified confidence level.}
\label{fig:ablation_plot}
\end{figure}

Figure \ref{fig:ablation_plot} is an ROC plot for all configurations producing confidence scores. The true positive and false positives reflect the inclusion of predictions with higher confidence, and the diameter of points is proportional to the count for each confidence label. 
When we compare the confidence-only system to its "conventional" variant, we notice that the performance for level 5 shifts dramatically to the 0.5 random-choice line, encompassing a greater share of false positives and a smaller portion of true positives.
We also see that \textit{no guess} provides the best AUC performance of all systems at level 5, but it presents no meaningful alternation with 4 (or any other confidence level.)

\section{Discussion}

This study shows that GPT-4 provides increasingly better output as it is given more opportunity to characterize and explain its response. 
This is more effective when the structure avoids complicating the logical structure of the task.
For example, we interpret the absolute 0.111 F1 gain from using our default confidence scoring scheme instead of the \textit{conventional} one as indicating that GPT-4 performs better when fewer logical operations are required. 
The task already probes the presence or absence of events of the given type; 
we maintain consistency by allowing the LLM to provide confidence in the \textbf{presence} of the event. 
The conventional approach requires an additional level of indirection, by asking for a characterization of the LLMs response, not the input data. 

If we also urge GPT-4 to guess and explain itself when uncertain, we get useful confidence estimations and a modest improvement in recall. 
Without these appeals, GPT-4's event extraction output is rather similar:
it continues to speculate, 
but the confidence scores no longer indicate less certain outputs.
We observe that GPT-4 is capable of directly indicating its level of confidence, but it needs to be given explicit license to speculate along with ample means to provide its stance to its response. 
GPT-4 has clearly been subject to significant examination and public scrutiny;
it is not an open-source effort, but
we speculate that \approachname bypasses features imposed onto GPT-4 to guard against embarrassing output. 
Gaining confidence scores with a useful AUC significantly increases the utility of models such as GPT-4 in developing and extending modest annotation resources for tasks such as event detection.

\section{Future Work} \label{sec:future_work}
Though modest in scope, this study presents a promising technique for eliciting useful confidence judgments from GPT-4 while improving F1 in the few-shot setting.
Various additional lines of research would expand our understanding of value of this approach. 
First, additional baselines would be helpful. 
Evaluating a BERT-like model on the present version of the \better \basic event detection task, both few-shot and fine-tuned, would facilitate the assessment of the accuracy of \approachname. 
A baseline for confidence estimation using a white-box technique, such as the use of the log-probabilities from GPT-4 for vertical pipes in output would help contextualize our reported AUC values.

Second, we did not explore the issue of \textit{calibration error}: many systems express confidence as probability of correctness, and analyses examine the statistical gap between these figures and observed accuracy rates. 
Instead, we only requested confidence scores 1-5, and we did not attempt to associate each of these scores with specific precision rates. 
We would also like to explore the use of \textit{verbalized} confidence \cite{tian-etal-2023-just}, which could easily be applied to the present task.

Other potential lines of study include the application of \approachname to LLMs other than GPT-4 and tasks beyond our variant of \better. We observed that prediction in the \textbf{presence} of a label was associated with better F1 and AUC than confidence in the \textbf{correctness} of the response, regardless of its content. Extending this logic to some NLP tasks could be challenging, but we would like to better understand the scope of applicability and relevance of this technique.

\section{Conclusion}
This study presents \approachname, which combines a set of prompting techniques for effectively performing the event detection task with GPT-4 in the few-shot setting. 
More importantly, we provide a strategy for exposing GPT-4's capacity to provide useful confidence scores.
This crucially depends upon urging the model to speculate and explain: 
simply providing the opportunity to explain is not enough. 
We suspect that GPT-4 has been engineered to default to indicating high certainty, 
and our appeal lifts this "hold" on the model's functionality.
However, the design and breadth of opportunities to explain also impact performance.
Eliminating explanations and yes/no questions lowers F1 and AUC.
We interpret this, in part, as GPT-4 being designed to avoid "traps": 
it indicates more of its weaknesses when encouraged to do so.
The venues for characterizing uncertainty need to be ample but also well suited to the task presented to the model, 
as shown with the degradation employing the "conventional" confidence approach.

\section{Limitations}
Section \ref{sec:future_work} described various limits of this study which could be addressed in later efforts. We also note that we only used GPT-4 and only examined annotation of English texts. 

\section*{Ethics Statement}
This system enhances F1 and confidence estimation, but many errors remain: users cannot assume that system output is accurate, when marked with high confidence.

\section*{Acknowledgements}
This material is based upon work supported by the US Air Force under Contract No. FA8750-22-C-0511. Any opinions, findings and conclusions or recommendations expressed in this material are those of the author(s) and do not necessarily reflect the views of the US Air Force. 
We also benefited from the feedback and support of Steve Minton, Goran Muric, and Michael Ross.

\bibliography{anthology,custom}
\bibliographystyle{acl_natbib}

\appendix

\end{document}